%% file: draft_main.tex
\documentclass[11pt]{article}

\usepackage{subfigure}
\usepackage{hhline}
\usepackage[space]{grffile}
\usepackage{multirow}

\usepackage{mathrsfs}
\usepackage{fullpage}
\usepackage[margin = 1in]{geometry}
\usepackage{times}
\usepackage{latexsym}
\makeatletter
\def\ps@headings{
\def\@oddhead{\mbox{}\scriptsize\rightmark \hfil \thepage}
\def\@evenhead{\scriptsize\thepage \hfil \leftmark\mbox{}}
\def\@oddfoot{}
\def\@evenfoot{}}
\makeatother

\usepackage[utf8]{inputenc} % allow utf-8 input
\usepackage[T1]{fontenc}    % use 8-bit T1 fonts
\usepackage{url}            % simple URL typesetting
\usepackage{booktabs}       % professional-quality tables
\usepackage{amsfonts}       % blackboard math symbols
\usepackage{nicefrac}       % compact symbols for 1/2, etc.
\usepackage{times}
\usepackage{graphicx}
\usepackage{epstopdf}
\usepackage{epsfig}
\usepackage{natbib}

\usepackage{latexsym}

\makeatletter
\def\ps@headings{%
\def\@oddhead{\mbox{}\scriptsize\rightmark \hfil \thepage}%
\def\@evenhead{\scriptsize\thepage \hfil \leftmark\mbox{}}%
\def\@oddfoot{}%
\def\@evenfoot{}}
\makeatother

\usepackage{wrapfig}
\makeatletter
\def\runningfoot{\def\@runningfoot{}}
\def\firstfoot{\def\@firstfoot{}}
\makeatother

\usepackage{amsmath}
\usepackage{amssymb}
\usepackage{amsthm}
\usepackage{amscd}
\usepackage{amsfonts}
\usepackage{graphicx}%
\usepackage{fancyhdr}
\usepackage{color}
\usepackage{algpseudocode}
\usepackage{algorithm}
\usepackage{enumitem}
\usepackage{bbm}

\def\BState{\State\hskip-\ALG@thistlm}

\makeatletter  % Add this!!!!
 \renewcommand{\ALG@name}{Mechanism} %doesnt work
\makeatother   % Add this also
\newtheorem{theorem}{Theorem}
\newtheorem{lemma}{Lemma}
\newtheorem{proposition}[theorem]{Proposition}
\newtheorem{definition}{Definition}
\usepackage{hyperref}
\newcommand{\PP}{\mathbb P}
\newcommand{\EE}{\mathbb E}
\newcommand{\Et}{\mathcal E}
\newcommand{\Exp}{\text{exp}}
\newcommand{\Pb}{\mathbb P}
\newcommand{\E}{\mathcal E}

\makeatletter
\def\@copyrightspace{\relax}
\makeatother

\newcommand{\squishlist}{
\begin{list}{{{\small{$\bullet$}}}}
{\setlength{\itemsep}{1pt}      \setlength{\parsep}{1pt}
\setlength{\topsep}{1pt}       \setlength{\partopsep}{0pt}
\setlength{\leftmargin}{1em} \setlength{\labelwidth}{1em}
\setlength{\labelsep}{0.5em} } }
\newcommand{\squishend}{  \end{list}  }

\ifodd 1
\newcommand{\com}[1]{\textbf{\color{red}(COMMENT: #1)}} %comment of the text
\newcommand{\clar}[1]{\textbf{\color{green}(NEED CLARIFICATION: #1)}}
\newcommand{\response}[1]{\textbf{\color{magenta}(RESPONSE: #1)}} %response to comment
\else

\newcommand{\com}[1]{}
\newcommand{\clar}[1]{}
\newcommand{\response}[1]{}
\fi
\newcommand{\RNum}[1]{\uppercase\expandafter{\romannumeral #1\relax}}
\newcommand{\argmin}{\text{argmin}}

\newcommand{\margdef}[1]{}

\begin{document}

%\twocolumn[

\title{Online Learning Using Only Peer Prediction}

\author{ Yang Liu\footnote{yangliu@ucsc.edu, Computer Science and Engineering, University of California, Santa Cruz} ~and~ David P. Helmbold\footnote{dph@soe.ucsc.edu, Computer Science and Engineering, University of California, Santa Cruz}}

\date{}

\maketitle

%\address{ yangliu@ucsc.edu\\Computer Science and Engineering\\ University of California, Santa Cruz \And  dph@soe.ucsc.edu\\Computer Science and Engineering\\University of California, Santa Cruz  } 
%]
\begin{abstract}
This paper considers a variant of the classical online learning problem with expert predictions. 
Our model's differences and challenges are due to lacking any direct feedback on the loss each expert incurs at each time step $t$. 
We propose an approach that uses peer prediction and identify conditions where it succeeds. 
Our techniques revolve around a carefully designed peer score function $s()$ that scores experts' predictions based on the peer consensus. We show a sufficient condition, that we call \emph{peer calibration}, under which standard online learning algorithms using loss feedback computed by the carefully crafted $s()$ have bounded regret with respect to the unrevealed ground truth values.  
We then demonstrate how suitable $s()$ functions can be derived for different assumptions and models.
\end{abstract}

\input{introduction}

\input{peer}

\input{sufficient}
\input{main-square}

\input{unknown}
\input{hetero}

\section{Concluding remarks}

In this paper, we developed a framework for online learning problems where peer assessment is the only feedback.  
We derived appropriate peer-score functions that can be used as proxies for the experts' losses
and showed they result in low-regret algorithms.
These score functions are more sophisticated than simply using the majority opinion as an artificial label.
With this lower level of feedback, additional assumptions are needed. To a certain degree, our solution provides a solution template for self-supervised online learning under different assumptions.

One direction for future work is to see if our assumption on a gap between best and next best experts
(with respect to the $g()$ function) can be removed or weakened.
Another direction for further work involves seeing how the peer-prediction framework can be extended to additional and more subtle relationships between the consensus artificial labels and the true labels, or even bypassing the need for artificial labels entirely. We'd also like to apply our methods to real-world datasets.

\section*{Acknowledgement}
Yang Liu would like to thank Yiling Chen for helpful and inspiring early discussions on this problem. This work is partially funded by
the Defense Advanced Research Projects Agency (DARPA) and Space and Naval Warfare Systems Center Pacific (SSC Pacific) under Contract No. N66001-19-C-4014. The views and conclusions contained herein
are those of the authors and should not be interpreted as necessarily representing the official policies, either
expressed or implied, of DARPA, SSC Pacific or the U.S. Government. The U.S. Government is authorized to reproduce and distribute reprints for governmental purposes notwithstanding any
copyright annotation therein.

\bibliographystyle{plainnat} 
\bibliography{myref,myref-yang,references,noise_learning,library}
\newpage
\onecolumn

\input{appendix}

\end{document}

%% file: introduction.tex
\section{Introduction}

Consider the following online expert selection problem: at discretized time steps $t=1,2,...,T$, each of $N$ experts will form a forecast about a binary event $y_t \in \{1~(\text{happening}), 0~(\text{not happening})\}$. 
Let's denote expert $i$'s prediction of how likely $y_t=1$ will happen at time $t$ as $p_i(t) \in [0,1]$. In the classical online learning setting, after each round $t$ (at time $t^+$), $y_t$ is observed and each expert incurs a loss $\underline{\ell_{i,t} := \ell(p_i(t),y_t)}$ according to a given loss function $\ell$, which can be the squared loss, a 0-1 loss, or some other loss function. 
The best expert is defined as the one whose predictions minimize the total losses in hindsight:
$
\underline{a^* = \argmin_i \sum_{t=1}^T \ell_{i,t}}.
$ 
At each round $t$, the algorithm selects an expert (often using randomization) and follows its prediction, denote the selected expert as \underline{$a(t)$}. 
To lighten the notation, we denote the prediction made according to selection $a(t)$,  i.e.~$p_{a(t)}(t)$, as  $p_a(t)$. 
 The algorithm's performance is typically evaluated using the following definition of regret:
\begin{align}
R_T := \EE \left[ \sum_{t=1}^T \ell_{a(t),t} \right] - \sum_{t=1}^T \ell_{a^*,t} \label{regret:def}
\end{align}
where the expectation is with respect to the algorithm's internal randomization, and the goal is to guarantee small regret $R_T$ (e.g.~sub-linear in $T$).

In several natural applications of online learning, neither the ground-truth $y_t$ values nor the true losses may be immediately available. 
One example is the hiring junior faculty candidates by committee in a large department.
Which faculty candidates will develop into superstars will only become apparent 
later in the faculty members' careers, and many offers must be issued
before this information becomes available.
Our setting involves taking into account the  opinions of experts (committee members) based on the particulars of the applicants at the time of hiring. Similarly school admission and other selection committees are also applications of our setting. Our goal is to identify and follow the best of these experts using a peer prediction method, where we will purely rely on predictions made by peer experts to identify proxies of the true losses. 
This setting also finds applications in other domains:
\squishlist
\item Crowdsourcing:  follow the best labelers, or learn how to best aggregate their advice, without ever knowing the ground truth labels.
\item Long-term forecasting: use predictions from experts made long before the outcomes are realized, update forecasters' weights in advance, and make better predictions.
This could correspond to the experts making all of their predictions at time 0.
\item Limited access to ground truth:  even when there is limited access to some ground truth values,
peer prediction allows more efficient use of this limited information.
\squishend

We study the situation where all (or sometimes most) of the $y_t$'s are unavailable, 
and thus the $\ell_{i,t}$'s cannot be directly computed. 
The goal is still to have small regret $R_T$ with respect to the (now unseen) $y_t$'s as defined in Eqn.~(\ref{regret:def}).
Our model is even more extreme than typical bandit problems: we do not even get the $\ell_{i,t}$ losses for the chosen experts.

With this paucity of feedback we must relax the adversarial setting typical in online learning models through some additional assumptions.
Instead of the unavailable true losses, we construct \emph{peer-score} functions, using peer prediction, to estimate the goodness of the experts' predictions.
A natural requirement is that the consensus of the expert's predictions is somehow correlated with the true outcomes.
We enforce this by requiring that both the original loss function and the peer-score function be ``calibrated'' by compatible divergence functions\footnote{Divergence functions are like distance functions but the triangle inequality may not hold, for instance in Bregman divergences.}.
Our analysis also needs a small gap between the best and second-best scoring experts under the peer-score function. Note that even though we require the consensus to be correlated with the ground truth, this correlation can be a weak one. Our work focuses on selecting the best experts instead of performing the optimal aggregation - in practice, a small committee of the best experts can often outperform the crowd consensus \citep{tetlock2016superforecasting,goldstein2014wisdom,sig15}. 

Our peer-score functions do not simply take the majority prediction as a proxy label:
they explicitly adjust for biases in the majority opinion.
This enables us to bound the regret when 
standard online algorithms are run using 
these peer scores as proxy losses.

\paragraph{Contributions and Outline:} 
Our contributions include formalizing a peer prediction framework to study online learning problems without ground truth feedback.  
This framework is developed in Section~\ref{sec:peer},
and involves relating the peer scores to the ground-truth losses through their calibrating functions.
A second contribution is formalizing conditions on the peer scoring that guarantees
any good online algorithm using the peer scoring will (w.h.p.) have  good regret with respect to the unseen labels (Theorem~\ref{main:convergence}).  
In addition, we derive suitable peer scoring functions for the square-loss with 
a methodology that generalizes to
other calibrated and bounded loss functions in Section~\ref{sec:realization}. This methodology assumes that the peer reference answers are related to the true labels
through a known i.i.d. noise rate.  
Our third contribution is relaxing this assumption, providing bounds for known asymmetric error rates and 
when the noise rate is unknown, but a converging estimate of it is available (also in Section~\ref{sec:realization}).
We then show how such a converging estimate of the i.i.d. error rate can be efficiently produced from limited access to ground truth (\ref{limitgt}) or even using just the expert's predictions (\ref{nogt}).
Finally, we examine time-varying error rates in Section~\ref{sec:hetero} and show how a competitive-style regret bound can be derived for that case. Our results can also be viewed as an effort to achieve self-supervision in online learning.  All proofs can be found in the Appendix. 
 %The proof of Theorem~\ref{main:convergence} is at the end of the main section, and most other proofs are in the supplementary material.

\paragraph{Related work:}
As a well-established research area, it is impossible to do a thorough survey on online learning in limited space, and we refer readers to \citep{cesa2006prediction} for a textbook treatment.
 Learning results can be categorized based on the types of feedback the problem admits, including: full feedback \citep{littlestone1994weighted,Cesa-Bianchi1997,arora2012multiplicative}, bandit feedback \citep{auer2002finite}, partial feedback \citep{mannor2011bandits}, graph feedback \citep{alon2015online}, among many others. 
Our results complement the online learning literature via introducing a solution framework that has no feedback but uses assumptions on peer predictions. 
The idea of using peer predictions has appeared in the peer prediction literature \citep{Prelec:2004,MRZ:2005,witkowski2012robust,radanovic2013,dasgupta2013crowdsourced,2016arXiv160303151S,kong2016equilibrium,radanovic2016incentives}. 
Peer prediction functions have the following nice property that experts' scores will be minimized if the event is happening with exactly their reported/forecasted probability. 
Our work is also relevant to the literature of learning with noisy data \citep{angluin1988learning,cesa2011online,natarajan2013learning,van2015learning,scott2015rate}. 
The ideas are also tied to establishing calibrated surrogate losses that are robust to label noise. 
However, knowledge of the noise rates are often assumed to be known. 
We provide fixes when such a priori knowledge is absent.

Some of our example applications resemble delayed feedback settings, which have been studied previously 
(e.g.~\citep{mesterharm2005line,joulani2013online,thune2019nonstochastic}). 
Although our paper makes stronger assumptions on the experts' predictions, the resulting bounds hold even if the feedback never arrives.

%% file: peer.tex
\section{Peer Prediction Framework and Main Result}

\label{sec:peer}

After stating the prediction model, we define calibrating functions $f()$ for the original loss function $\ell()$
and $g()$ for the peer-scores $s()$.  
We then define the compatibility of  $f()$ and $g()$ needed for our main result, and state our main theorem
bounding the regret when appropriate peer-scores are used.

\subsection{Preliminaries}

\paragraph{Prediction model}
At each round $t$ the following happens:
\squishlist
\item Nature selects an unknown outcome distribution $p_t$.
\item Outcome $y_t$ for the occurrence of event $t$ is drawn with $y_t \sim p_t$.
\item Each expert $i$ predicts a probability $p_i(t)$ that event $t$ occurs, possibly based on the context of the current and previous events.
\item The algorithm selects, perhaps with the aid of randomization, 
an expert $a(t)$ and predicts with $p_{a(t)}(t)$, based only on the experts' current and past predictions. 
\squishend
Although one can consider the $p_t$ and $p_i(t)$ values as generated adversarially, the purpose of the paper is to examine what reasonable assumptions on the $p_i(t)$ values lead to successful learning with peer feedback.

As mentioned earlier, our goal is to minimize 
$$
R_T = \EE \left[ \sum_{t=1}^T \ell_{a(t),t} \right] - \sum_{t=1}^T \ell_{a^*,t}.
$$ 
We will also use $\underline{L_i :=  \sum_{t=1}^T \ell_{i,t}}$
for the total loss of expert $i$ with respect to the ground truth $y_t$.
Since we appeal to martingale inequalities, the $p_t$'s must depend only on the previous trials.

\paragraph{Peer prediction}
Instead of using $y_t$ which remains largely unavailable, 
the algorithm uses a reference answer $\hat{y}_{t}$ to evaluate each expert $i$'s prediction. 
In short, $\hat{y}_{t} \in \{0,1\}$ is some aggregation of the experts' predictions:
$
\underline{\hat{y}_{t} :=\mathcal A(\{p_i(t)\}_{i =1}^N)}, 
$
where $\mathcal A(\cdot)$ maps the predictions of all experts to a single estimated label. 
For instance, $\mathcal A(\cdot)$ can be taken as the majority votes of the thresholded experts' predictions,  or the ``most likely'' $y$-value found by comparing $\prod_{i=1}^N p_i(t)$ with $\prod_{i=1}^N (1-p_i(t))$. 

We will call $\hat{y}_{t}$ a peer \emph{reference answer}. 
Then a \emph{peer-score} function
$
\underline{s_{i,t}:=s(p_i(t), \hat{y}_{t})}
$
 is used as a proxy for the loss of expert $i$'s prediction. 
 We aim to study what $s()$, when combined with standard online learning algorithms, 
 guarantees a small regret $R_T$ (with respect to the unseen $y_t$). 
 Of course, when $s$ or $\hat{y}_{t}$ is not properly designed, 
 the peer-scores may not characterize the true performance of each expert. 
For instance, simply checking each prediction against the majority opinion of the set of experts may not properly identify the best expert -- rather it will elect the ones who predict the majority opinion more. 
We will see later that suitable $s()$'s are more subtle.

\subsection{Loss calibration}
\begin{definition}
A loss function $\ell$ is \underline{$f$-calibrated} if 
$$
\mathbb E_{y \sim p}[\ell(p',y)] - \mathbb E_{y \sim p}[\ell(p,y)] = f(p', p),
$$
where $f()$ is a (non-negative) divergence function that measures the difference between $p$ and $p'$.
\end{definition}
If the loss is $f$-calibrated, then the second term $\mathbb E_{y \sim p}[\ell(p,y)]$ is the \emph{minimum} expected loss that can be achieved, and it corresponds to the loss of a genie who predicts with the true distribution of $y$. 
We now give an example of an $f$-calibrated loss.

\begin{lemma} \label{f:calibrate}
Squared loss $\ell(p_a(t),y) = (y-p_{a}(t))^2$ is calibrated with $f(p_a(t),p_t))=(p_t-p_a(t))^2$. 
\end{lemma}
Throughout this paper, we will use squared loss as the running example, but our results generalize to other bounded proper losses, thanks to the Savage representation \citep{Gneiting:07} (see Appendix). 
If $\ell$ is $f$-calibrated, we have the following:
\[
\sum_{t=1}^T \mathbb E_{y_t \sim p_t} \left[ \ell_{i,t} \right]  -  \sum_{t=1}^T \mathbb E_{y_t \sim p_t}\left[\ell(p_t,y_t) \right] = \sum_{t=1}^T f(p_i(t),p_t)
\]
The second term, 
$\sum_{t=1}^T \mathbb E_{y_t \sim p_t}\left[\ell(p_t,y_t) \right]$
corresponds to the best possible forecaster that predicts with the distributions used to draw the outcomes $y_t$.
Let $a^*_f$ be the best expert w.r.t. $f()$:  \margdef{{$a^*_f$}}
$
a^*_f = \argmin_i \sum_{t=1}^T f(p_i(t),p_t).
$ 

We'd like to argue that the best expert $a^*$ in hindsight should roughly (and with high probability) minimize $\sum_{t=1}^T f(p_i(t),p_t)$, due to the 
convergence of $\sum_{t=1}^T \ell_{i,t}$ and $\sum_{t=1}^T \ell(p_t,y_t)$. Define $\mathcal H_t$ as the information set of relevant history up to time $t$, including all earlier $y_{t'}$'s, and $p_i(t')$'s, $t' \leq t$. We will use the following martingale lemma:
\begin{lemma}\label{lemma:martingale}
Let $q(1), q(2), \ldots$ be a sequence of prediction distributions where each $q(t)$ depends only on $\mathcal H_{t-1}$ (and is thus conditionally independent of $y_t$), then 
$\ell_{t} := \sum_{\tau=1}^t \ell(q(\tau),y_\tau) - \sum_{\tau=1}^t \ell(p_\tau,y_\tau)-\sum_{\tau=1}^t f(q(\tau),p_\tau)$ formulates a martingale. 
\end{lemma}

The above lemma, together with the convergence properties of martingales, 
implies that, with high probability, the expert with the minimum sum of $f$ scores also has low loss with respect
to the true labels, so $L_{a^*_f} \approx L_{a^*}$. More precisely, the Hoeffding-Azuma inequality for martingales gives
the following bound for any $\Et_{mart}>0$:
\begin{align}
\Pb \biggl (\biggl| \sum_{\tau=1}^t \ell(q(\tau),y_\tau)& - \sum_{\tau=1}^t \ell(p_\tau,y_\tau)-\sum_{\tau=1}^t f(q(\tau),p_\tau) \biggr| \geq \Et_{mart} \biggr) \leq 2 \exp \biggl( -\frac{\Et_{mart}^2}{8t}\biggr) \label{HZ:f}
\end{align}
\begin{lemma}\label{af}
With prob. at least $1-2N \cdot \exp \biggl( -\frac{\Et^2_{mart}}{32T}\biggr)$, 
we have $L_{a^*_f} \leq L_{a^*} + \Et_{mart}$.
\end{lemma}
%Lemma~\ref{af}'s proof is very similar to the one for Lemma \ref{cali:g}, see the proof of that lemma in the Appendix.

Recall that $p$ is the probability that $y=1$, and let $\hat p$ be the probability that the reference feedback $\hat y = 1$.
 We define calibration for the peer-score function as follows.
\begin{definition}
A peer-score function $s()$ is \underline{$g$-calibrated} if 
\begin{align}
\mathbb E_{\hat{y} \sim \hat p}[s(p',\hat{y})] - \mathbb E_{\hat{y} \sim \hat  p}[s(p,\hat{y})] = g (p', p)
\end{align}
where $g()$ is a divergence function measuring the difference between $p$ and $p'$ in the context of $\hat p$.
\end{definition}

Since $\hat p$ appears on the left-hand-side, $g()$ will in general depend on $\hat p$ and it could be treated as an additional argument.  
However, we assume that $\hat p$ is the same function of $p$ over all rounds, \margdef{{$g()$}} and thus
are able to suppress this dependency.   
This is the case if, for example, each $\hat y_t$ is an i.i.d. $\eta$-perturbation of $y_t$ so $\Pb (\hat y_t \neq y_t) = \eta$.
Later in the paper we will consider alternative ways of generating the reference labels,
but the analysis will implicitly use a function $g()$ whose $\hat p_t$ probabilities are a fixed function of $p_t$. 

Let $a^*_g$ be the best expert with respect to $g$ and the $p_t$ values:  \margdef{{$a^*_g$}}
$
\underline{a^*_g = \argmin_i \sum_{t=1}^T g(p_i(t),p_t)},
$~ and let $a^*_{peer}$ be the best expert with respect to $s()$:   \margdef{{$a^*_{peer}$}}
$
\underline{a^*_{peer} = \argmin_i \sum_{t=1}^T s_{i,t}}.
$ 
Consider running a ``no regret" online learning algorithm over the experts using $s(p_i(t), \hat y_t) $ for the expert's losses.  
The guarantee of the online learning algorithm bounds the following regret \citep{cesa2006prediction}:  \margdef{{$R^{peer}_T$}}
\begin{align}
R^{peer}_T := \EE\left[\sum_{t=1}^T s_{a(t),t}\right] - \sum_{t=1}^T s_{a^*_{peer},t}  
\end{align}
Our goal is to use this bound on $R^{peer}_T$ to obtain bounds on $R_T$. 
As before, the Hoeffding-Azuma inequality for martingales  easily gives the following bound for any $r>0$, where   \margdef{{$\sigma_g$}}
$$
\sigma_g \geq |s(q(\tau),\hat{y}_\tau) - s(p_\tau,\hat{y}_\tau)- g(q(\tau),p_\tau)|
$$ 
bounds the magnitude of the changes to the random variables:  
\begin{align}
\Pb \biggl ( \biggl|\sum_{\tau=1}^t s(q(\tau),\hat y_\tau) &- \sum_{\tau=1}^t s(p_\tau, \hat y_\tau)-\sum_{\tau=1}^t g(q(\tau),p_\tau) \biggr| \geq r \biggr) \leq 2\text{exp}\biggl( -\frac{r^2}{2\sigma^2_g \cdot t}\biggr)\label{conv:S}
\end{align}

%Let  \margdef{{$\Delta^g_i, \Delta^g_{\min}$}}
%$
%\underline{\Delta^g_i = \sum_{t=1}^T s_{i,t} - \sum_{t=1}^T s_{a^*_{peer},t}},~\underline{\Delta^g_{\min} := \min_{i \neq a^*_g} \Delta^g_i}$. Applying Eqn. (\ref{conv:S}) yields the following:
%\begin{lemma}\label{cali:g}
%%\end{lemma}
%For instance when $\Delta^g_{\min} \geq \sqrt{8\sigma_g^2 T\cdot \log T}$, the above probability term becomes $1-O(\frac{N}{T})$.

It is important to realize that although the true loss $\ell$ is needed (counterfactually) to evaluate for the ultimate regret, 
and the peer-score function $s()$ is needed to run the algorithm,
the corresponding calibrating functions $f()$ and $g()$ are used only for the analysis.  

We now come to the key definition of the paper.  
This definition establishes a connection between the true losses $\ell()$ and the peer-scores $s()$ 
through a relationship between their 
calibrating functions $f()$ and $g()$.   
Very informally, it says that if the algorithm's predictions and the predictions of the best expert with respect to $g()$ have related $g$-divergences, then the 
algorithm's predictions and the predictions of the best expert with respect to $f()$ 
have somewhat similar $f$-divergences.  
This is what will allow us to move from peer-score regrets to regrets on the true losses.
It may be more surprising that peer scoring functions with the needed property can be constructed for natural situations than that this connection leads to good regret bounds. 

\begin{definition}
We call $g$  ``\underline{$\psi$-compatible} with $f$'' if there exists an invertible, increasing, and convex function $\psi$ with $\psi(0) = 0$ such that for all $p_t$
\begin{align*}
f(p_a(t),&p_t)-
f(p_{a^*_f}(t),p_t) \leq \psi^{-1}\biggl(g(p_a(t),p_t)-
g(p_{a^*_{g}}(t),p_t)  \biggr)
\end{align*}
\end{definition}

This definition is essentially the $\psi$-transform in supervised learning \citep{bartlett2006convexity}. 
Compatibility gives a very strong relationship between $f$ and $g$. 
In particular, If $f$ and $g$ are $\psi$-compatible, then immediately:

\paragraph{Fact 1} $\psi^{-1}$ is concave and increasing, and $\psi^{-1}(0) = 0$.

This peer calibration leads us to the following propositions (proven in the Appendix):

\begin{proposition}\label{sum:calibrate}
If  $g$ is $\psi$-compatible with $f$, then:
\begin{align*}
&\sum_{t=1}^T f(p_a(t),p_t) - \sum_{t=1}^T f(p_{a^*_f}(t),p_t) \leq T \cdot \psi^{-1} \biggl( \frac{\sum_{t=1}^T g(p_a(t),p_t) - \sum_{t=1}^T g(p_{a^*_g}(t),p_t)}{T}\biggr)
\end{align*}
\end{proposition}

\begin{proposition}\label{g=f}
If  $g$ is $\psi$-compatible with $f$, then: there exist $a^*_g,~a^*_f$ such that
$a^*_g = a^*_{f}$. %(assuming $a^*_{f}$ is unique).
\end{proposition}

%% file: sufficient.tex
\subsection{Peer calibration is sufficient}
\label{sec:sufficient}
We are now ready to sketch the proof of our main theorem: that learning from the peer-score $s()$ losses leads to low-regret with respect to the $\ell()$ losses on the unseen ground-truth $y_t$ values.  
The proof proceeds by first observing that the
peer-scored loss of the algorithm is at most the peer-scored loss of $a^*_{peer}$ plus the algorithm's expected regret bound, which we write as $\underline{\Et_{online} (T,N)} \in O(\sqrt{T\ln N})$.  

We use the martingale relationship between the $s()$ losses and its $g()$ calibration and the Hoeffding-Azuma
inequality to show that the $s()$ losses for $a^*_{peer}$ and the predictions used by the algorithm
are closely related to their calibrating $g()$ values.
We denote the tolerable gap with 
$\underline{\Et_{mart}(\delta, \sigma_g, T) = \sqrt{2\sigma_g^2 \cdot T \cdot \ln \frac{2}{\delta}}}$
(recall that $\sigma_g$ is the scale parameter for martingale sequence),
this guarantees that each is within the gap with probability $1-\delta$. 

The optimalities of $a^*_{peer}$ and $a^*_g$ for $s(\cdot)$ and $g(\cdot)$ respectively return us a fact that the total sum of $g()$ values for the algorithm's predictions are 
within $\Et_{online}(T,N) + 2 \Et_{mart} (\delta, \sigma_g,T)$ of the total for the optimizing $a^*_g$
with probability at least $1-2\delta$. The compatibility between $f()$ and $g()$  ensures that $a^*_f = a^*_g$, 
so $a^*_f$ is also likely to incur the same $g(\cdot)$ as $a^*_{peer}$.   
This compatibility allows us to use $\psi^{-1}$ to convert average per-trial closeness wrt $g()$ 
into closeness wrt $f()$.  
%Assuming a gap of $\Delta^g_{\min}$ between $a^*_g$ and the next best expert with respect to $g()$
%allows us to again use the martingale relationship to show that $a^*_g = a^*_{peer}$ 
%with probability at least $1-\delta_g$ where  $\underline{\delta_g := 2N \cdot \text{exp}\biggl( -\frac{(\Delta^g_{\min})^2}{8\sigma^2_g  T}\biggr)}$. 

Another pair of martingale inequalities show the $\ell()$ actual losses with respect to the ground-truth $y_t$'s
are closely related to the calibrated $f()$ functions for $a^*_f$ and the $a(t)$ predictions used by the algorithm.
Gaps of $\Et_{mart} (\delta, 2, T) = \sqrt{2 \cdot 2^2 \cdot T \cdot \ln \frac{2}{\delta} }$ are needed to show that each is within the gap with probability $1-\delta$.  
Adding these gaps to the regret bound (and subtracting another $2 \delta$ from the confidence) gives the following theorem:

\begin{theorem}\label{main:convergence}
If  $g$ is $\psi$-compatible with $f$, then with probability at least $1-4\delta$,
\begin{align}
R_T \leq T \cdot \psi^{-1} \biggl (&\frac{2\Et_{mart}(\delta, \sigma_g, T) + \Et_{online}(T,N) }{T}\biggr )+ 2\Et_{mart}(\delta, 2, T)
\end{align}
\end{theorem}

%Here %$\Et_{online}()$ is the regret bound of the on-line learning algorithm, 
%the $\Et_{mart}()$ functions ensure that the martingale bounds hold with probability $1-\delta$,
% and $1-\delta_g$ is the probability that $a^*_g = a^*_{peer}$ from Lemma~\ref{cali:g}.

%% file: main-square.tex
\section{Application to Square Loss}
\label{sec:realization}

When the loss and peer-score functions are calibrated with compatible functions, a small regret with respect to the unseen $y$ outcomes results using the peer-score for the experts' losses.    
In this section, we derive a suitable peer-score $s()$ and compatible calibrating $g()$ for the square loss.

We start by assuming each reference $\hat y_t$ is a perturbed version of $y_t$ with a symmetric (label independent) and homogeneous (time independent) perturbation probability $\eta$:
\[
\mathbb P(\hat{y}_{t} \neq y_{t}) = \eta, ~\text{with $\eta < 0.5$ }
\]
i.e.~$\hat y$ is better than random guessing. Although this homogeneous error rate assumption looks restrictive, it is weaker than the common one in the inference literature in crowdsourcing where all agents' error rates are assumed to be homogeneous. 
In practice, an aggregated reference answer is relatively more stable across different tasks, especially when the population is large.  

We initially assume $\eta$ is known, but then extend the analysis to 
the non-symmetric case and when only an approximation to $\eta$ is available.
Further extensions are in the following section.

\subsection{A peer prediction function and its regret}
Take $\ell$ as the squared loss: 
$\ell(p_a(t),y) = (y-p_{a}(t))^2$. From Lemma \ref{f:calibrate},
$
f(p_a(t),p_t)  = (p_t-p_a(t))^2
$ calibrates $\ell()$, therefore:
\begin{align*}
&\mathbb E_{y_t \sim p_t}\biggl [\sum_{t=1}^T \ell_{i,t} \biggr]  - \mathbb E_{y_t \sim p_t}\biggl [\sum_{t=1}^T \ell(p_t,y_t) \biggr] = \sum_{t=1}^T f(p_i(t),p_t) =\sum_{t=1}^T (p_t-p_i(t))^2
\end{align*}
Denote the true probability of $\hat{y}_t = 1$ with \margdef{{$\hat p_t$}}
 $\hat{p}_t$. 
Simple algebra shows that
\begin{align}
\underline{\hat{p}_t:=\mathbb P(\hat{y}_{t} =1)} &= \mathbb P(\hat{y}_{t} =1|y_t=1) \mathbb P(y_t=1) +  \mathbb P(\hat{y}_{t} =1|y_t=0) \mathbb P(y_t=0) \nonumber \\
&= (1-\eta) \cdot p_t + \eta\cdot (1-p_t)= (1-2\eta) \cdot p_t + \eta.\label{flip}
\end{align}
This observation 
enables us to prove the following lemma with a bit of simple algebra.
First we define:
\[
F(\eta,p_t) := - \eta(1-\eta)(1-2p_t)^2 + 2\eta \cdot p_t^2-2\eta \cdot p_t+\eta
\]
which is independent of $i$.

\begin{lemma}\label{lemma:noisye}
For expert $i = 1,...,N$ and time $1 \leq t \leq T$:
\begin{align*}
& \mathbb E_{\hat{y}_t \sim \hat{p}_t} \bigl [ \ell(p_i(t), \hat{y}_t )\bigr]  -  \mathbb E_{\hat{y}_t \sim \hat{p}_t}\bigl [ \ell (p_t,\hat{y}_t) \bigr] = (1-2\eta)f(p_i(t), p_t) - 2\eta\cdot p_i(t)(1-p_i(t))- F(\eta,p_t).
\end{align*}
\end{lemma}
The above lemma inspires us to design the following peer-score function $s(\cdot)$
by first cancelling the $p_i(t)(1-p_i(t))$ terms in $\ell(p_i(t),\hat{y}_t)$ and then observing  that 
$(1-2\eta)f(p_i(t), p_t) -F(\eta, p_t) $ is compatible with $f$ since $F(\eta, p_t)$ is invariant across all experts. 

\begin{theorem}\label{thm:peer}
If the peer-score function and $g(p_i(t), p_t)$ are:
\begin{align}
 &s_{i,t} :=  \ell(p_i(t),\hat{y}_t) + 2\eta\cdot p_i(t)(1-p_i(t)) \label{def:S}, \\ \label{E:sdef}
 & g\left(p_i(t), p_t\right) := (1-2 \eta) (p_t - p_i(t))^2 - F(\eta, p_t) - 2\eta   p_t(1-p_t),~
\end{align}
then $s()$ is $g()$-calibrated and $g$ is  $\psi^{-1}(x)= x/ (1-2\eta)$-compatible with $f()$.
\end{theorem}

Therefore Theorem~\ref{main:convergence} gives the following regret bound, which holds with probability $1-4\delta $:
\begin{small}
\begin{align*}
 R_T \leq& T \cdot \psi^{-1} \biggl (\frac{2\Et_{mart}(\delta, \sigma_g, T) + \Et_{online}(T,N) }{T}\biggr) + 2\Et_{mart}(\delta, 2, T)\\
 =& \frac{2\Et_{mart}(\delta, \sigma_g, T) + \Et_{online}(T,N) }{1-2\eta}+2\Et_{mart}(\delta, 2, T)
\end{align*}
\end{small}
where $\sigma_g = \max\{4 + \max F(\eta, p_t), 2-\min F(\eta, p_t)\}$. A couple of remarks follow:
\squishlist
\item The above bound assumes $\eta < 0.5$ and diverges as the $\hat y$ become uninformative
($\eta \rightarrow 1/2$).
\item Theorem~\ref{thm:peer}'s  peer-score construction 
can be generalized to other calibrated loss functions $\ell$ using the Savage representation of proper scoring rules/calibrated loss functions \citep{Gneiting:07} (see Appendix).
\squishend

\input{twocoin}

\subsection{Using estimated noise rates}

In practice, the error rate of $\hat y$ is unknown a priori. Before considering the learning of error rates, we generalize Theorem~\ref{main:convergence} and show how using an estimate $\hat \eta_t$ for
$\eta_t = \Pb (\hat y_t \neq y_t)$ affects the regret bounds.
Suppose the peer-score becomes (adapted from Eqn. (\ref{def:S})) 
$$
s_{i,t} := \ell(p_i(t),\hat{y}_t) + 2\hat{\eta}_t \cdot p_i(t)(1-p_i(t)),
$$ 
with $\hat \eta_t$ replacing $\eta$, 
and we have a bound $| \hat \eta_t - \eta | \leq \epsilon_t$
then we get the following.

\begin{theorem}\label{reg:error}
Suppose noisy estimates $\hat{\eta}_t$ replace the true noise rate $\eta$ in Eqn. (\ref{E:sdef})
where each $| \hat \eta_t - \eta | \leq \epsilon_t$, and the algorithm uses the resulting peer-scores.
Then Theorems~\ref{main:convergence} and~\ref{thm:peer} imply, 
with probability at least $1-4\delta$
 \begin {align*}
R_T \leq & \frac{2\Et_{mart}(\delta, \sigma_g, T) + \Et_{online}(T,N) + \sum_{t=1}^{T} \epsilon_t }{1-2\eta} + 2\Et_{mart} (\delta, 2, T)
\end{align*}
where $\sigma_g = \max\{4 + \max F(\eta, p_t), 2-\min F(\eta, p_t)\}$. 
\end{theorem}

%% file: twocoin.tex
\subsection{Asymmetric error rate}

We now relax the assumption of symmetric label noise: for known $\eta_0$ and $\eta_1$, let
\[
\mathbb P(\hat{y}_{t} =1 | y_{t}=0) = \eta_0, \qquad \mathbb P(\hat{y}_{t} =0 | y_{t}=1) = \eta_1,
\]
with $\eta_0 + \eta_1 < 1$ (better than random guessing) \cite{liu2017machine}. A more general approach relates to learning with noisy data  
\citep{natarajan2013learning,scott2015rate,menon2015learning,van2015learning}, where the goal is to design a surrogate loss function that calibrates the true losses in the presence of label biases. 
For instance, one such $s$ can be defined as follows:
\begin{align}
s(p_a(t), \hat{y}_t) =& (1-\eta_{1-\hat{y}_t})\ell(p_a(t), \hat{y}_t) -\eta_{\hat{y}_t} \ell(p_a(t), 1-\hat{y}_t) \label{eqn:sl}
\end{align}
Then we have 
\begin{lemma}[\cite{natarajan2013learning}]
For each time $t$, 
$
\mathbb E[s(p_a(t), \hat{y}_t) ] = (1-\eta_0-\eta_1) \cdot \mathbb E[\ell(p_a(t),y_t)].
$
\end{lemma}
Following above lemma immediately we will have
\begin{proposition}
$s()$ defined in $Eqn. (\ref{eqn:sl})$ is $g$-calibrated 
where $g():=(1-\eta_0-\eta_1) f()$ is $\psi$-compatible with $f$ for $\psi^{-1}(x) = x/ (1-\eta_0-\eta_1)$.
\end{proposition}

Therefore we establish the following regret bound from Theorem~\ref{main:convergence}
\begin{align*}
 \frac{2\Et_{mart}(\delta, \sigma_g, T) + \Et_{online}(T,N)}{1-\eta_0-\eta_1}+2\Et_{mart}(\delta, 2, T).
\end{align*}

Estimating the two error rates $\eta_0$ and $\eta_1$ is generally a harder task than estimating a single error rate, especially when the errors may vary over time (a challenge addressed in Section \ref{sec:learn} and \ref{sec:hetero}). 

\paragraph{Mapping to a class-independent error rate setting} In light of above discussion, we propose an approach to map the asymmetric error rate case to a symmetric one. At each time $t$ the trial is ``flipped'' 
with probability $1/2$.  
When a trial is ``flipped'' we use outcome $\tilde y_t := 1-y_t$ and
flipped predictions $\tilde p_i(t) := 1-p_i(t)$, so $\hat y_t$ is also flipped.
After flipping, $\hat{y}_t$ has the nice property:
\begin{lemma}\label{flip:symmetric}
$\hat{y}_t$ has class-independent error rates w.r.t. $\tilde y_t$.
\end{lemma}
This result allows us to focus on the class-independent error rate setting.

%% file: unknown.tex
\section{Approximating the error rates}
\label{sec:learn}
Here we extend the analysis to when the error rates of the reference answers are \emph{unknown} (Section \ref{limitgt} and \ref{nogt}) and \emph{heterogeneous} across time (Section \ref{sec:hetero}), expanding the applicability of our results.

\subsection{Limited access to ground truth}\label{limitgt}
Suppose the error rate $\eta = \Pb(\hat y_t \neq y_t)$ of the reference answer is 
homogeneous but is unknown a priori. We start with an easier setting where we occasionally get the $y_t$ ground truth feedback. 
When the ground truth is only revealed according to a certain probability, 
the standard way to handle information revealed according to a certain probability is 
to apply importance weighting to observed losses $\ell$. 
We show that limited access to ground truth can be better used to estimate
the $\hat y_t$ error rate, rather than learning the losses directly. 
Suppose, at each time $t$, the ground truth label becomes available with probability $p^*$.   
We apply importance weighting to estimate the error rate $\eta$ as follows:  
\[
   \hat{\mathbbm 1}(\hat{y}_t,y_t)= 
\begin{cases}
    \frac{\mathbbm 1(\hat{y}_t = y_t)}{p^*}, & \text{if ground truth becomes available }\\
    0,              & \text{otherwise}
\end{cases}
\]
then we estimate $\eta$ as follows at step $t$:
$
\hat{\eta}_t := \frac{\sum_{n=1}^t \hat{\mathbbm 1}(\hat{y}_t,y_t)}{t}
$. 
The expectation $\mathbb E[\hat{\eta}_t] = \eta$, next we show this estimation costs another $O(\frac{\sqrt{T \cdot \ln \frac{2}{\delta}}}{p^*})$ regret term in $\psi^{-1}(\frac{\cdot}{T})$ with probability at least $1-\delta$ (using Theorem~\ref{reg:error}). 

By the ``maximal" version of Hoeffding-Azuma inequality we know ( $\hat{\mathbbm 1}(\hat{y}_t,y_t)$ forms a martingale is again due to the martingale nature of $y_t$s)
\begin{align}
\PP\biggl(\max_{t \leq T}|\sum_{n=1}^t \hat{\mathbbm 1}(\hat{y}_t,y_t) - \eta \cdot t| > \epsilon\biggr) \leq 2\Exp\left(\frac{-2\epsilon^2}{t \cdot (\frac{1}{p^*})^2}\right)
\end{align}
Let $\epsilon = \sqrt{\frac{t}{2(p^*)^2} \ln \frac{2}{\delta}}$, we have with probability at most $\delta$ that: 
$
\biggl |\sum_{n=1}^t \hat{\mathbbm 1}(\hat{y}_t,y_t) - \eta \cdot t \biggr | > \sqrt{\frac{t}{2(p^*)^2} \ln \frac{2}{\delta}}$. Therefore 
\begin{align}
&|\hat{\eta}_t-\eta| = |\frac{\sum_{n=1}^t \hat{\mathbbm 1}(\hat{y}_t,y_t)}{t} - \frac{\eta \cdot t}{t}|\leq  \frac{\sqrt{\ln \frac{2}{\delta}}}{p^* \sqrt{2t}},~\forall t
\end{align}
with probability at least $1-\delta$. According to Theorem~\ref{reg:error}, this will introduce another regret term:
\[
\sum_{t=1}^T \epsilon_t = \sum_{t=1}^T \frac{\sqrt{\ln \frac{2}{\delta}}}{p^* \sqrt{2t}}= O\bigl(\frac{\sqrt{T \cdot \ln \frac{2}{\delta}}}{p^*}\bigr)
\]

Estimating a single error rate allows the $\frac{1}{p^*}$ term to be independent of the number of experts (as opposed to the typical $\frac{\sqrt{T  \ln (N / \delta) }}{p^*}$ regret \citep{cesa2006prediction}). 

\subsection{No access to ground truth}\label{nogt}
The task of estimating the error rate $\eta$ is much harder when there is no ground truth information available. We propose the following method to estimate it:
\squishlist
\item Randomly partition the experts into two groups, namely groups $A, B$. 
Denote the aggregated reference answers within each group as $\hat{y}_{A,t}$ and $\hat{y}_{B,t}$ respectively.  
\item Denote the error rates for $\hat{y}_{A,t}$ and $\hat{y}_{B,t}$ as $\eta_A,\eta_B$ respectively. 
Assume $\eta_A,\eta_B<0.5$, the error rates stay constant over time, and they are conditionally independent given the ground truth $y_t$: 
$\Pb(\hat{y}_{A,t},\hat{y}_{B,t}|y_t)=\Pb(\hat{y}_{A,t}|y_t)\Pb(\hat{y}_{B,t}|y_t)$.
\squishend
We leverage the comparison between the two groups. 
Define $c_{1,t}, c_{2,t}, c_{3,t}$ as the following (unknown) parameters estimatable without $y_t$s:
\begin{align*}
c_{1,t} &= \frac{\sum_{\tau=1}^t \PP(\hat{y}_{A,\tau} = 1)}{t},\\
c_{2,t} &=  \frac{\sum_{\tau=1}^t \PP(\hat{y}_{B,\tau}=1)}{t},\\
c_{3,t} &= \frac{\sum_{\tau=1}^t \PP(\hat{y}_{A,\tau} = \hat{y}_{B,\tau}=1)}{t}
\end{align*}
We have the following theorem:
\begin{theorem}
Rates $\eta_A,\eta_B < 1/2$ are uniquely characterized by the following three equations:
\begin{align*}
&P_{0,t} \cdot \eta_A + (1-P_{0,t}) (1-\eta_A) = c_{1,t},\\
&P_{0,t} \cdot \eta_B + (1-P_{0,t}) (1-\eta_B) = c_{2,t}\\
&P_{0,t} \cdot \eta_A \cdot \eta_B + (1-P_{0,t}) (1-\eta_A)(1-\eta_B) = c_{3,t},~
\end{align*}
where $P_{0,t} =\frac{\sum_{\tau=1}^t \mathbbm{1}(y_\tau = 0)}{t}$, $\eta_A$, and $\eta_B$ are the unknowns.
\label{learnable}
\end{theorem}
Parameters $c_{1,t},c_{2,t},c_{3,t}$ can be empirically estimated along the way, providing estimates for $\eta_A, \eta_B$ via solving the equations. 
Then we can set $\hat{y}_t$ as either $\hat{y}_{A,t}$ or $\hat{y}_{B,t}$, and use the estimated $\hat{\eta}_A,\hat{\eta}_B$ correspondingly. 

Denote the estimation of $\eta_A, \eta_B$ at time $t$ as $\hat{\eta}_{A,t}, \hat{\eta}_{B,t}$ respectively using estimates of $c_{1,t},c_{2,t},c_{3,t}$. A finer degree analysis also gives us:
\begin{theorem}\label{error:peer}
At $t$, w.p. $\geq 1-3\delta$, 
$
|\hat{\eta}_{A,t}-\eta_A | \leq O(\sqrt{\frac{\ln \frac{2}{\delta}}{2t}}), ~|\hat{\eta}_{B,t}-\eta_B| \leq O(\sqrt{\frac{\ln \frac{2}{\delta}}{2t}}),
$ when $P_0$ is bounded away from 0.5.\footnote{When $P_0$ is close to 0.5, the first and second equations presented in the estimation equations in Theorem \ref{learnable} can uniquely determine $\eta_A,\eta_B$ separately. }
\end{theorem}
This will translate to a $O(\sqrt{\frac{\ln (6/\delta)}{2t}})$ regret bound for $\eta_A$ and $\eta_B$ with probability at least $1-\delta$, which incurs an additional $\sum_{t=1}^T O(\sqrt{\frac{\ln (6/\delta)}{2t}}) = O(\sqrt{T\cdot \ln \frac{6}{\delta}})$ regret, per Theorem~\ref{reg:error}. 

%The Appendix shows using these estimates costs only an $O(\sqrt{T\cdot \ln \frac{6}{\delta}})$ additional regret term with probability at least $1-\delta$.

%% file: hetero.tex
\subsection{Heterogeneous error rates}
\label{sec:hetero}
Now we consider a setting where the error rates, now denoted $\eta_t<0.5$,  change. 
The challenge is the previous techniques lead to minimizing a term like (according to Lemma \ref{lemma:noisye} and Theorem \ref{thm:peer}):
$$
 \sum_{t=1}^T (1-2\eta_t)f(p_a(t), p_t) \sim \sum_{t=1}^T (1-2\eta_t) (\ell_{a,t} - \ell(p_t,y_t))
$$
instead of the constant $1-2\eta$ coefficient, which enables compatible calibration. 
Our previous error estimation procedure estimates the average error rate
instead of treating each $\eta_t$ separately.

Inspired by the uniform noise case, if the $\eta_t$s can be made similar enough, then peer calibration techniques can give bounds even in the heterogeneous case.
We use the following flipping based mechanism to reduce the heterogeneity: randomly flip the peer reference answer with probability $\hat{p}$:
\[
\tilde{y}_t := 
\begin{cases}
    \hat{y}_t, & \text{w.p. } 1-\hat{p}\\
    1-\hat{y}_t,              &\text{w.p. } \hat{p}
\end{cases}
\]
and use this newly flipped $\tilde{y}_t $ as our peer reference outcome. 
With this flipping, the error rate \underline{$\tilde \eta_t$} for reference answer $\tilde{y}_t$ becomes: 
$
{\tilde{\eta}_t = \eta_t (1-\hat{p}) + (1-\eta_t) \hat{p}}.
$
This implies that for any two times $t_1,t_2$ we have
$
|\tilde{\eta}_{t_1}-\tilde{\eta}_{t_2}| := (1-2\hat{p}) |\eta_{t_1}-\eta_{t_2}|.
$ 
Let \underline{$\tilde{\eta}$ be the average $\frac{\sum_{t=1}^T \tilde{\eta}_t}{T}$}, implying
$$
|\tilde{\eta}_t - \tilde{\eta}| \leq (1-2\hat{p}) \max_{t_1,t_2} |\eta_{t_1}-\eta_{t_2}|
.$$
As $\hat{p} \rightarrow 0.5$, the slack in this inequality becomes arbitrarily small, and the different error rates at different $t$ become similar (homogeneous).
Thus a properly chosen $\hat{p}$ can make $|\eta_t - \tilde{\eta}|$ small enough
to exploit the similarity between the $f()$ and $g()$ functions almost as if they were compatible.

With this flipping, we can estimate $\eta_t$ as the average error rate up to time $t$ using methods from 
Sections~\ref{limitgt} and~\ref{nogt} 
 for use in the peer-scores, denoting as $\hat{\eta}_t$. And then let $$
s_{i,t} = \ell(p_i(t), \tilde{y}_t) + \hat{\eta}_t \cdot p_i(t) \cdot (1-p_i(t)).
$$ We now focus on binary expert predictions where $p_i(t) \in \{0,1\}$. Note all our previous results hold for the binary prediction case as $p_i(t)$s can be interpreted as with probability 0 or 1.  
For the competitive ratio $c_{\textsf{comp}}(\alpha):= \alpha\bigl(\frac{1}{1-2\max_t \tilde{\eta}_t}+1\bigr)$, 
we have:
\begin{theorem}\label{reg:hetero}
For any $\alpha=2+\epsilon$ ($\epsilon > 0$), there exists a $0<\hat{p}<1/2$ (bounded away from 0.5) such that,
with probability at least $1-\delta-\delta_g$,  the above process's regret $R_T$ is bounded as follows:
\begin{align*}
R_T \leq&  \frac{\Et_{mart}(\frac{\delta}{2N}, 2, T)+\Et_{mart}(\frac{\delta}{2N}, \sigma_g, T)+\Et_{online}(T,N)}{1-2\max_t \tilde{\eta}_t} +c_{\textsf{comp}}(\alpha) \cdot L_{a^*}.~
\end{align*}
\end{theorem}
Thus we achieve a competitive ratio w.r.t.~the optimal loss, up to an additional sub-linear term. Note $\max_t \tilde{\eta}_t$ is bounded away from 0.5 if both $\hat{p}$ and $\eta_t$s are.

%% file: appendix.tex
%%appendix

\section*{Appendix}

We provide the missing proofs.

\subsection*{Summary of key notations}

\begin{table}[h]
\caption{Notation table}

\centerline{
\begin{tabular}{| c | l | }
\hline
Symbol 	& Meaning \\
\hline\hline
$a^*$ 	&  least-loss expert $ \argmin_{1\leq i \leq N} \sum_{t=1}^T \ell_{i,t}$ \\
$a^*_f$	& best expert wrt loss-calibrated $f$: $\argmin_i \sum_{t=1}^T f(p_i(t),p_t)$ \\
$a^*_g$	& best expert wrt peer-calibrated $g$: $\argmin_i \sum_{t=1}^T g(p_i(t),p_t)$ \\
$a^*_{peer}$	& best expert wrt peer-prediction loss $s$: $\argmin_i \sum_{t=1}^T s_{i,t}$ \\
$a(t)$	& algorithm's distribution used at time $t$ \\
$\mathcal A(\{p_j(t)\}_{j}), \hat{y}_{i}$ & "reference ground truth" from experts, same as $\hat{y}_{t}$ (discrete prediction)\\
$\hat{p}_t$ & probability $\hat{y}_{i}=1$.\\
$f(p',p)$		& divergence function, loss $f$-calibrated if $\mathbb E_{y \sim p}[\ell(p',y)] - \mathbb E_{y \sim p}[\ell(p,y)] = f(p', p)$ \\
$g(p',p)$		& peer loss calibration divergence function, \\
$\mathcal H_t$	& relevant history up to time $t$, all earlier $y_{t'}$'s, and $p_i(t')$'s, $t' \leq t$. \\
$\ell(p_i(t),y_t)$	& loss function, taking expert's prob distribution and outcome \\
$\ell_{i,t}$	& shorthand for $\ell(p_i(t),y_t)$ \\
$\hat{\ell}_{i,t}$	& shorthand for $\ell(p_i(t),\hat{y}_t)$ \\
$N$ 		& number of experts \\
$p_a (t)$	& algorithm's random prediction drawn from $a(t)$ \\
$p_i (t)$ 	& prob. distr. of expert $i$ at trial $t$ \\
$p_t$ 	& prob. distr. for event/outcome $t$, so $y_t | \mathcal H_{t-1} \sim p_t$ \\
$R_T$	& total regret, $\sum_{t=1}^T \ell_{a(t),t} - \sum_{t=1}^T \ell_{a^*,t}$  \\
$R^{peer}_T$ 	& total algorithm regret wrt peer loss, $\sum_{t=1}^T s_{a(t),t} - \sum_{t=1}^T s_{a^*_{peer},t} $ \\
$s_{i,t}$, $s()$ 	& $s(p_i(t), \hat{y}_{t})$, peer prediction loss function, calibrated by $g()$ \\
$T$  		& number of trials/timesteps \\
$y_t$ 	& 0-1 label at time $t$ \\
$\psi() $	& ``peer calibrated loss funct.",  bounds $f$ difference in terms of $g$ differences:   \\
		& \qquad $ f(p_a(t),p_t)- f(p_{a^*_f}(t),p_t) \leq \psi^{-1}\biggl(g(p_a(t),p_t)- g(p_{a^*_{g}}(t),p_t)  \biggr) $ \\
%$\Delta_i$ & the gap between the true accumulated losses incurred by the $i$-th expert and the best expert. \\
%$\Delta_{\min}$ & the minimum of $\Delta_i$, besides the best expert. \\
%$\Delta^g_i$ & the gap between the accumulated $s$ losses incurred by the $i$-th expert and the best expert. \\
%$\Delta^g_{\min}$ & the minimum of $\Delta^g_i$, besides the best expert. \\
$\sigma_g$ & bound of the magnitude of the martingale difference incurred by $s()$\\
\hline
\end{tabular}
}
\end{table}

\subsection*{Proof for Lemma \ref{f:calibrate}}
\begin{proof}
\begin{align*}
&f(p_a(t),p_t) \\
=& \mathbb E_{y_t \sim p_t}[\ell(p_a(t), y_t)] - \mathbb E_{y_t \sim p_t}[\ell(p_t, y_t)]  \\
=& p_t \cdot (1-p_a(t))^2 + (1-p_t)\cdot p_a(t)^2 - (p_t \cdot (1-p_t)^2 + (1-p_t)\cdot p_t^2)\\
=&(p_t-p_a(t))^2 + p_t(1-p_t)- ((p_t-p_t)^2 + p_t(1-p_t))\\
=&(p_t-p_a(t))^2
\end{align*}
\end{proof}

\subsection*{Proof for Lemma \ref{lemma:martingale}}

\begin{proof}
\begin{align*}
\mathbb E \left[ \ell_t | \mathcal H_{t-1} \right] &= \mathbb E \left[\sum_{\tau=1}^t \ell(q(\tau), y_\tau) - \sum_{\tau=1}^t \ell(p_\tau,y_\tau)-\sum_{\tau=1}^t f(q(\tau),p_\tau)| \mathcal H_{t-1} \right]  \\
&=\sum_{\tau=1}^{t-1} \ell(q(\tau), y_\tau) - \sum_{\tau=1}^{t-1} \ell(p_\tau,y_\tau) - \sum_{\tau=1}^{t-1} f(q(\tau),p_\tau)\\
&~~~~~+ \mathbb E_{y_t \sim p_t}[\ell(q(t), y_t) -\ell(p_t,y_t) | \mathcal H_{t-1}]- f(q(t),p_t)\\
&=\ell_{t-1},
\end{align*}
where the last equality is by conditional independence and $f$-calibration. 
\end{proof}

\subsection*{Proof for Lemma \ref{af}}

\begin{proof}
Via union bound and applying Eqn. \ref{HZ:f} we know that with probability at least 
\[
1-2N \cdot \exp \biggl( -\frac{\Et_{mart}^2}{32 T}\biggr)
\]
we have 
\begin{align}
\biggl| \sum_{t=1}^T \ell(p_{a^*_f}(t),p_t)& - \sum_{t=1}^T \ell(p_t, y_t)-\sum_{t=1}^t f({a^*_f},p_\tau) \biggr| < \frac{\Et_{mart}}{2} 
\end{align}
and
\begin{align}
\biggl| \sum_{t=1}^T \ell(p_{a^*}(t),y_\tau)& - \sum_{t=1}^T \ell(p_t,y_t)-\sum_{t=1}^T f(p_{a^*},p_t) \biggr| < \frac{\Et_{mart}}{2} 
\end{align}
Therefore 
\begin{align*}
L_{a^*_f} &\leq  \sum_{t=1}^T \ell(p_t, y_t)+\sum_{t=1}^t f({a^*_f},p_\tau)+ \frac{\Et_{mart}}{2}\\
&\leq  \sum_{t=1}^T \ell(p_t, y_t)+\sum_{t=1}^t f({a^*},p_\tau)+ \frac{\Et_{mart}}{2}\\
&\leq  \sum_{t=1}^T \ell(p_{a^*}(t),y_\tau) +  \frac{\Et_{mart}}{2}+  \frac{\Et_{mart}}{2}\\
&= L){a^*} + \Et_{mart}.
\end{align*}

%Then 
%\begin{align*}
%\sum_{t=1}^T g(p_i(t),p_t) - \sum_{t=1}^T g(p_{a^*}(t),p_t)
%\end{align*}
%\[
%|\sum_{t=1}^T s(p_i(t),y_t) - \sum_{t=1}^T s(p_t,y_t)-\sum_{t=1}^T g(p_i(t),p_t)| < \frac{\Delta^g_{\min}}{2}, ~\forall i
%\]
%Therefore for $i \neq a^*_{peer}$ we have
%\begin{align*}
%& \sum_{t=1}^T g(p_i(t),p_t) - \sum_{t=1}^T g(p_{a^*}(t),p_t) \\
%> &\bigl(\sum_{t=1}^T s(p_i(t),y_t) - \sum_{t=1}^T s(p_t,y_t) - \frac{\Delta^g_{\min}}{2}\bigr) - \bigl( \sum_{t=1}^T s(p_{a^*}(t),y_t) - \sum_{t=1}^T s(p_t,y_t) + \frac{\Delta^g_{\min}}{2}\bigr)\\
%=&\sum_{t=1}^T s(p_i(t),y_t)-\sum_{t=1}^T s(p_{a^*}(t),y_t) -\Delta^g_{\min} \geq 0
%\end{align*}
%This proves that $a^*_{peer} = a^*_g$. 

\end{proof}
\subsection*{Proof for Proposition \ref{sum:calibrate}}

\begin{proof}
\begin{align*}
&\sum_{t=1}^T f(p_a(t),p_t) - \sum_{t=1}^T f(p_{a^*_f}(t),p_t) \\
\leq & \sum_{t=1}^T \psi^{-1}\bigl(g(p_a(t),p_t)-
g(p_{a^*_g}(t),p_t) \\
= & T\cdot \frac{\sum_{t=1}^T \psi^{-1}\bigl(g(p_a(t),p_t)-
g(p_{a^*_g}(t),p_t)}{T}\\
\leq & T \cdot \psi^{-1}\biggl(  \frac{\sum_{t=1}^T g(p_a(t),p_t) - \sum_{t=1}^T g(p_{a^*_g}(t),p_t)}{T}\biggr)
\end{align*}
where the last inequality is due to concavity of $\psi^{-1}$.
\end{proof}
\subsection*{Proof for Proposition \ref{g=f}}

\begin{proof}
\begin{align}
\sum_{t=1}^T f(p_{a^*_g}(t),p_t) - \sum_{t=1}^T f(p_{a^*_f}(t),p_t) \leq& T \cdot \psi^{-1} \left(\frac{ \sum_{t=1}^T g(p_{a^*_g}(t),p_t) - \sum_{t=1}^T g(p_{a^*_g}(t),p_t)}{T}\right) = 0\\
\Rightarrow &\sum_{t=1}^T f(p_{a^*_g}(t),p_t) \leq \sum_{t=1}^T f(p_{a^*_f}(t),p_t) \\
\Rightarrow &\sum_{t=1}^T f(p_{a^*_g}(t),p_t) = \sum_{t=1}^T f(p_{a^*_f}(t),p_t) \label{equality}
\end{align}
where Eqn. (\ref{equality}) is due to the optimality of $a^*_f$.
This concludes the proof.
\end{proof}
\subsection*{Proof of Theorem \ref{main:convergence}}

\begin{proof}
We list the key steps in the proof. 

\noindent \textbf{Step 1} Using Martingale inequality we know
\begin{align*}
&\sum_{t=1}^T s_{a(t),t} - \sum_{t=1}^T s_{p_t,t}  \rightarrow \sum_{t=1}^T g(p_a(t),p_t),\\
&\sum_{t=1}^T \ell_{a(t),t} - \sum_{t=1}^T \ell_{p_t,t}  \rightarrow \sum_{t=1}^T f(p_a(t),p_t)
\end{align*}
In particular, from Eqn. (\ref{conv:S}), with probability at least $1-2\delta$, 
the following holds:
%\begin{small}
\begin{align*}  %\label{e:genT}
\biggl |\sum_{t=1}^T s_{a(t),t}& - \sum_{t=1}^T s_{p_t,t}  - \sum_{t=1}^T g(p_a(t),p_t)\biggr |  \leq \Et_{mart}(\delta, \sigma_g, T)\\
\biggl |\sum_{t=1}^T s_{a^*_{peer},t}& - \sum_{t=1}^T s_{p_t,t} - \sum_{t=1}^T g(p_{a^*_{peer}}(t),p_t)\biggr | \leq \Et_{mart}(\delta, \sigma_g, T)
\end{align*}
%\end{small}
Similarly with probability at least $1-2\delta$,  
%\begin{small}
\begin{align*} % \label{e:genT}
\biggl |\sum_{t=1}^T \ell_{a(t),t} &- \sum_{t=1}^T \ell_{p_t,t} - \sum_{t=1}^T f(p_a(t),p_t)\biggr |  \leq \Et_{mart}(\delta, 2, T)\\
\biggl |\sum_{t=1}^T \ell_{a^*_{peer},t}& - \sum_{t=1}^T \ell_{p_t,t} - \sum_{t=1}^T f(p_{a^*_{peer}}(t),p_t)\biggr | \leq \Et_{mart}(\delta, 2, T)
\end{align*}
%\end{small}

\paragraph{Step 2} %Via Lemma \ref{cali:g}, we know with probability at least $1-\delta_g$ that $a^*_g = a^*_{peer}$. 
Using facts in Step 1, we know the following holds:% with probability at least $1-\delta_g$:
\begin{align*}
&\sum_{t=1}^T g(p_a(t),p_t)-\sum_{t=1}^T g(p_{a^*_g}(t),p_t) \\
\leq &\sum_{t=1}^T g(p_a(t),p_t)-\sum_{t=1}^T s_{a(t),t} + \sum_{t=1}^T s_{p_t,t}  \\
&~~+  \sum_{t=1}^T s_{a^*_{g},t}-\sum_{t=1}^T s_{p_t,t}  -\sum_{t=1}^T g(p_{a^*_g}(t),p_t) \\
&~~+  \sum_{t=1}^T s_{a(t),t} - \sum_{t=1}^T s_{a^*_{peer},t}\\
\leq & 2\Et_{mart}(\delta, \sigma_g, T) +\sum_{t=1}^T s_{a(t),t} - \sum_{t=1}^T s_{a^*_{peer},t}% \Et_{online}(T,N) 
\end{align*}
The first inequality is because $\sum_{t=1}^T s_{a^*_{peer},t} \leq  \sum_{t=1}^T s_{a^*_g,t}$ (optimality of $a^*_{peer}$).

\paragraph{Step 3} By Proposition \ref{sum:calibrate} we know  \begin{small}
\begin{align*}
& \sum_{t=1}^T f(p_a(t),p_t) - \sum_{t=1}^T f(p_{a^*_f}(t),p_t) \\
\leq & T \cdot \psi^{-1} \biggl( \frac{\sum_{t=1}^T g(p_a(t),p_t) - \sum_{t=1}^T g(p_{a^*_g}(t),p_t)}{T}\biggr)\\
\leq & T \cdot \psi^{-1} \biggl (\frac{2\Et_{mart}(\delta, \sigma_g, T) + \sum_{t=1}^T s_{a(t),t} - \sum_{t=1}^T s_{a^*_{peer},t}}{T}\biggr )
\end{align*}
\end{small}

\paragraph{Step 4} Then  
%\begin{small}
\begin{align*}
 &\sum_{t=1}^T \ell_{a(t),t} - \sum_{t=1}^T \ell_{a^*,t} \\
 =&(\sum_{t=1}^T \ell_{a(t),t} - \sum_{t=1}^T \ell_{p_t,t}) - (\sum_{t=1}^T \ell_{a^*,t} - \sum_{t=1}^T \ell_{p_t,t})\\
 \leq & \sum_{t=1}^T f(p_a(t),p_t) - \sum_{t=1}^T f(p_{a^*}(t),p_t) + 2\Et_{mart}(\delta, 2, T) \\
 \leq & \sum_{t=1}^T f(p_a(t),p_t) - \sum_{t=1}^T f(p_{a^*_f}(t),p_t)+ 2\Et_{mart}(\delta, 2, T) \\
 \leq & T \cdot \psi^{-1} \biggl (\frac{2\Et_{mart}(\delta, \sigma_g, T) + \sum_{t=1}^T s_{a(t),t} - \sum_{t=1}^T s_{a^*_{peer},t} }{T}\biggr )\\
 &~~~~ +2\Et_{mart}(\delta, 2, T)
\end{align*}
%\end{small}

\paragraph{Step 5} From the guarantee of running an online learning algorithm, we have
\begin{align}
\EE\left[\sum_{t=1}^T s_{a(t),t}\right]-\sum_{t=1}^T s_{a^*_{peer},t} \leq \Et_{online}(T,N) 
\end{align}
Further 
\begin{small}
\begin{align*}
 &\EE\left[\sum_{t=1}^T \ell_{a(t),t}\right] - \sum_{t=1}^T \ell_{a^*,t} \\
 \leq & T \cdot \EE\left[ \psi^{-1} \biggl (\frac{2\Et_{mart}(\delta, \sigma_g, T) + \sum_{t=1}^T s_{a(t),t} - \sum_{t=1}^T s_{a^*_{peer},t} }{T}\biggr )\right]\\
 &~~~~ +2\Et_{mart}(\delta, 2, T)\\
 \leq & T \cdot  \psi^{-1} \biggl (\frac{2\Et_{mart}(\delta, \sigma_g, T) +\EE\left[ \sum_{t=1}^T s_{a(t),t} - \sum_{t=1}^T s_{a^*_{peer},t}\right] }{T}\biggr )\\
 &~~~~ +2\Et_{mart}(\delta, 2, T)\quad \quad\quad\quad \text{(Concavity of $\psi^{-1}(\cdot)$)}\\
  \leq & T \cdot  \psi^{-1} \biggl (\frac{2\Et_{mart}(\delta, \sigma_g, T) +\Et_{online}(T,N)  }{T}\biggr )\\
 &~~~~ +2\Et_{mart}(\delta, 2, T).
 \end{align*}
 This completes the proof.
\end{small}
\end{proof}

\subsection*{Proof for Theorem \ref{thm:peer}}
\begin{proof}  
Denote by $\hat{\ell}_{i,t}:=\ell(p_i(t),\hat{y}_t)$. With 
$$
s(p_i(t), \hat y_t) = \hat \ell_{i,t} + 2\eta \cdot p_i(t) ( 1-p_i(t) ),
$$ 
we have
 \begin{align*}
& \mathbb E_{\hat y_t \sim \hat p_t} \bigl [ s_{i,t} \bigr]  -  \mathbb E_{ \hat y_t \sim \hat p_t } \bigl [s(p_t,\hat{y}_t) \bigr] \\
 =& (1-2\eta) (p_t-p_i(t))^2 -2\eta  p_i(t)(1-p_i(t)) -F(\eta,p_t)\\
&\qquad +2\eta p_i(t)(1-p_i(t)) -2\eta p_t(1-p_t)  ~~ \text{(using Lemma\ref{lemma:noisye}) } \\
=&(1-2\eta) \underbrace{(p_t-p_i(t))^2}_{f(p_i(t),p_t)} - F(\eta, p_t)-2\eta   p_t(1-p_t) \\
:=& g(p_i(t),p_t)
\end{align*}
Next, it is not hard to see that the minimizer $a^*_g$ over experts $i$ for 
\begin{align*}
\sum_{t=1}^T  g(p_i(t),p_t) &= (1-2\eta) \sum_{t=1}^T f(p_i(t),p_t) \\
&- \sum_{t=1}^T \left(G(\eta,p_t) + 2\eta   p_t(1-p_t) \right)
\end{align*}
is the same as the expert minimizing
$
\sum_{t=1}^T f(p_i(t),p_t)
$, 
as the former is simply an affine transform of the latter, so $a^*_g = a^*_f$. 

Now we show $g$ defined above is $\psi$-peer-calibrated 
with $\psi( \cdot) = \cdot / (1-2\eta)$. 
 \begin{align*}
&g(p_a(t),p_t) - g(p_{a^*_g}(t),p_t) \\
 =&  (1-2\eta) \bigl( (p_t-p_a(t))^2 -(p_t-p_{a^*_g}(t))^2 \bigr)\\
=&(1-2\eta) (f(p_a(t),p_t) - f(p_{a^*_f}(t),p_t) )
\end{align*}
\end{proof}

\subsection*{Proof for Lemma \ref{lemma:noisye}}

\begin{proof}

\begin{align*}
\mathbb E_{\hat{y}_t \sim \hat{p}_t} \bigl [ \ell(p_i(t), \hat{y}_t ) \bigr]  -  \mathbb E_{\hat{y}_t \sim \hat{p}_t}\bigl [ \ell(p_t,\hat{y}_t) \bigr]  
=& f(p_i(t),\hat{p}_t) 			& \text{($\ell$ is $f$-calibrated})\\
=&(\hat{p}_t -p_i(t))^2			& \text{(Lemma 1)} \\
=&\biggl ((1-2\eta) \cdot p_t + \eta-p_i(t)\biggr )^2 	& \text{assumption on $\hat p_t$} \\
=&\biggl ((p_t-p_i(t)) - \eta(2p_t-1)\biggr )^2 \\
\end{align*}
The above further derives as follows:
\begin{align*}
&\biggl ((p_t-p_i(t)) - \eta(2p_t-1)\biggr )^2\\
=&(p_t-p_i(t))^2 +\eta^2(2p_t-1)^2 - 2\eta \cdot (p_t-p_i(t))(2p_t-1)\\
=&(p_t-p_i(t))^2 +\eta^2(2p_t-1)^2 - 2\eta \cdot p_t(2p_t-1)+ 2\eta \cdot p_i(t) (2p_t-1)-2\eta p_t+\eta\\
=&(p_t-p_i(t))^2 - \eta(1-\eta)(1-2p_t)^2 + 4\eta \cdot p_i(t)p_t - 2\eta \cdot p_i(t)-2\eta p_t+\eta\\
=&(p_t-p_i(t))^2 - \eta(1-\eta)(1-2p_t)^2 - 2\eta \cdot ((p_i(t)-p_t)^2 -p_i(t)^2 - p_t^2)- 2\eta\cdot p_i(t)-2\eta p_t+\eta\\
=&(1-2\eta)(p_t-p_i(t))^2 - 2\eta\cdot p_i(t)(1-p_i(t))- F(\eta,p_t)
\end{align*}
where $F(\eta,p_t) := - \eta(1-\eta)(1-2p_t)^2 + 2\eta p_t^2-2\eta p_t+\eta$ which is independent of $p_i(t)$.
\end{proof}

\subsection*{Savage representation}
\label{sec:savage}
According to \cite{Gneiting:07}, savage representation states that for a strictly proper scoring function we have 
\[
S(p,y) = G(e^y) - \mathcal D_{G}(e^y,p),
\]
 where $e^y$ is an all-0 vector with only 1 for the component corresponding to outcome $y$. $ \mathcal D_{G}$ is the Bregman divergence w.r.t. $G$: \[  \mathcal D_{G}(e^y,p) = G(e^y) - G(p) - \bigtriangledown G(p) \cdot (e^y-p),\] where $\bigtriangledown G(p)$ is a sub-gradient of $G$.  For binary setting, we have 
 \begin{align*}
 S(p,1) &= G(p)+(1-p) G'(p)\\
 S(p,0) &= G(p)-p G'(p)
 \end{align*}
 Since a calibrated loss function corresponds to a strictly proper scoring function \cite{reid2011information,Gneiting:07}, the above representation is also true:
\[
\ell(p,y) = G(e^y) - G(p) - \bigtriangledown G(p) \cdot (e^y-p)-G(e^y) + \phi(y),
\]
where $\phi(y)$ is a term that is independent of $p$. Or in the binary case,
 \begin{align*}
 \ell(p,1) &= -G(p)-(1-p) G'(p)+\phi(1)\\
 \ell(p,0) &= -G(p)+p G'(p)+\phi(0)
 \end{align*}
 To summarize 
 \[
 \ell(p,y) =-G(p)-(1-p)^{y}p^{1-y} G'(p) + \phi(y).
 \]
Taking expectation of the loss evaluated at a noisy ground truth $\hat{y}$:
\begin{align*}
&\mathbb E[\ell(p,\hat{y})] = \mathbb E[-G(p) - (1-\eta)(1-p)^{y}p^{1-y} G'(p) \\
&~~~~~- \eta \cdot (1-p)^{1-y}p^{y} G'(p)] + \mathbb E[\phi(\hat{y})]\\
=&\mathbb E[-G(p) - (1-\eta)(1-p)^{y}p^{1-y} G'(p) \\
&~~~~~- \eta \cdot (1-(1-p)^{y}p^{1-y}G'(p)]+ \mathbb E[\phi(\hat{y})]\\
=&(1-2\eta)\mathbb E[(-G(p)-(1-p)^{y}p^{1-y} G'(p))] \\
&~~~~~+ 2\eta \cdot G(p) - \eta \cdot G'(p)+ \mathbb E[\phi(\hat{y})]\\
=&(1-2\eta)\mathbb E[\ell(p,y)] + 2\eta \cdot G(p) - \eta \cdot G'(p)+ \mathbb E[\phi(\hat{y})]
\end{align*}
Therefore we can always cancel $ 2\eta \cdot G(p) - \eta \cdot G'(p)$  by adding the term to $s()$ to return a compatible peer-score function.

\subsection*{Proof for Lemma \ref{flip:symmetric}}

\begin{proof}
\begin{align*}
&\Pb(\hat{y}_t=0|y^{\star}_t=1) = \Pb(\hat{y}_t=0,\mathcal E |y^{\star}_t=1) + \Pb(\hat{y}_t=0,~\bar{\E}|y^{\star}_t=1)\\
=&  \Pb(\hat{y}_t=0|y^{\star}_t=1, \E)\cdot \Pb(\E) +   \Pb(\hat{y}_t=0|y^{\star}_t=1, \bar{\E})\cdot \Pb(\bar{\E}) \\
=& \Pb(\hat{y}_t=1| y_t =0)\cdot \Pb(\E) +   \Pb(\hat{y}_t=0| y_t=1)\cdot \Pb(\bar{\E}) \\
=&\frac{\eta_1  + \eta_0}{2},
\end{align*}
where the second equality is due to the independence of the flipping $\E$. Similarly
\begin{align*}
&\Pb(\hat{y}_t=1|y^{\star}_t=0) = \Pb(\hat{y}_t=1,\mathcal E |y^{\star}_t=0) + \Pb(\hat{y}_t=1,~\bar{\E}|y^{\star}_t=0)\\
=&  \Pb(\hat{y}_t=1|y^{\star}_t=0, \E)\cdot \Pb(\E) +   \Pb(\hat{y}_t=1|y^{\star}_t=0, \bar{\E})\cdot \Pb(\bar{\E}) \\
=& \Pb(\hat{y}_t=0| y_t =1)\cdot \Pb(\E) +   \Pb(\hat{y}_t=1| y_t=0)\cdot \Pb(\bar{\E}) \\
=&\frac{\eta_1 + \eta_0}{2},
\end{align*}

\end{proof}

\subsection*{Proof for Theorem~\ref{reg:error}}
\begin{proof}
Estimating the noise rate with $\hat{\eta}_t$ leads to a noisy version
of the peer-score function $s()$ defined in Eqn.~(\ref{def:S}) via the term $2\eta p_i(t)(1-p_i(t))$.  
Since
\begin{align*}
2p_a(t)(1-p_a(t)) \leq 1/2, ~\forall p_a(t) \in [0,1],
\end{align*}
we have:
\begin{align*}
|2\hat{\eta}_t \cdot p_a(t)(1-p_a(t))-2\eta \cdot  p_a(t)(1-p_a(t))| \leq  |\hat{\eta}_t-\eta| / 2 \leq \epsilon_{t} / 2.
\end{align*}
Denote $\hat{s}()$ as the peer-score function defined using the estimates $\hat \eta$, and use it as the proxy loss for the online learning algorithm.  
Recalling that $\Et_{online}(T,N)$ is the online algorithm's regret bound:
\begin{align*}
\Et_{online}(T,N) &\geq \EE\left[ \sum_{t=1}^T \hat{s}_{a(t), t} \right] - \min_i \sum_{t=1}^T \hat{s}_{i, t}  \\
	& \geq \EE \left[ \sum_{t=1}^T \hat{s}_{a(t), t} \right] - \sum_{t=1}^T \hat{s}_{a^*_{peer}, t}  \\
	& \geq  \EE \left[ \sum_{t=1}^T {s}_{a(t), t} \right] - \sum_{t=1}^T {s}_{a^*_{peer}, t}    - \sum_t \epsilon_t .
\end{align*}

The rest of the regret analysis follows from Theorem~\ref{thm:peer} and 
the proof of Theorem~\ref{main:convergence}  with 
$\Et_{online}(T,N)$ replaced by  $\Et_{online}(T,N) + \sum_t \epsilon_t$.
\end{proof}

%\subsection*{Justification for the additional regret term in Section \ref{limitgt}}
%By the ``maximal" version of Hoeffding-Azuma inequality we know ( $\hat{\mathbbm 1}(\hat{y}_t,y_t)$ forms a martingale is again due to the martingale nature of $y_t$s)
%\begin{align}
%\PP\biggl(\max_{t \leq T}|\sum_{n=1}^t \hat{\mathbbm 1}(\hat{y}_t,y_t) - \eta \cdot t| > \epsilon\biggr) \leq 2\Exp(\frac{-2\epsilon^2}{t \cdot (\frac{1}{p^*})^2})
%\end{align}
%Let $\epsilon = \sqrt{\frac{t}{2(p^*)^2} \ln \frac{2}{\delta}}$, we have with probability at most $\delta$ that: 
%$
%\biggl |\sum_{n=1}^t \hat{\mathbbm 1}(\hat{y}_t,y_t) - \eta \cdot t \biggr | > \sqrt{\frac{t}{2(p^*)^2} \ln \frac{2}{\delta}}$. Therefore 
%\begin{align}
%&|\hat{\eta}_t-\eta| = |\frac{\sum_{n=1}^t \hat{\mathbbm 1}(\hat{y}_t,y_t)}{t} - \frac{\eta \cdot t}{t}|\leq  \frac{\sqrt{\ln \frac{2}{\delta}}}{p^* \sqrt{2t}},~\forall t
%\end{align}
%with probability at least $1-\delta$. According to Theorem~\ref{reg:error}, this will introduce another regret term:
%\[
%\sum_{t=1}^T \epsilon_t = \sum_{t=1}^T \frac{\sqrt{\ln \frac{2}{\delta}}}{p^* \sqrt{2t}}= O\bigl(\frac{\sqrt{T \cdot \ln \frac{2}{\delta}}}{p^*}\bigr)
%\]

\subsection*{Proof for Theorem \ref{learnable}}
\begin{proof}
We prove the uniqueness of the solution $\eta_A,\eta_B<0.5$ in the next three steps.

\paragraph{Step 1} The true parameters satisfy the system of equations:

For the first equation, when $y_t = 0$, $\Pb(\hat{y}_{A,t} = 1) = \eta_A$ and when $y_t = 1$, $\Pb(\hat{y}_{A,t} = 1) = 1-\eta_A$. This is also true for the second equation. For the third, when $y_t = 0$, $ \Pb(\hat{y}_{A,t} = \hat{y}_{B,t}=1) = \eta_A \cdot \eta_B$, and $y_t = 1$, $ \Pb(\hat{y}_{A,t} = \hat{y}_{B,t}=1) = (1-\eta_A) \cdot (1-\eta_B)$.

\paragraph{Step 2} There exits at most two solutions, via reducing the solutions to the solutions of a quadratic equation. First 
\begin{align*}
\eta_A = \frac{c_{1,t} - (1-P_{0,t})}{2P_{0,t}-1}, ~\eta_B = \frac{c_{2,t} - (1-P_{0,t})}{2P_{0,t}-1},\\
1-\eta_A = \frac{P_{0,t} - 2 - c_{1,t} }{2P_{0,t}-1}, ~1-\eta_B = \frac{P_{0,t} - 2 - c_{2,t}}{2P_{0,t}-1}
\end{align*}
(Dropping the $t$-subscript for ease of presentation) Plugging into the third equation we have 
\begin{align}\label{P0}
P_0 (c_1 - (1-P_0))(c_2 - (1-P_0)) + (1-P_0)(P_0 - 2 - c_1 )(P_0 - 2 - c_2 ) = c_3 (2P_0-1)^2
\end{align}
which is a quadratic equation of $P_0$, which says there exist at most two solutions:
\[
d_1 P^2_0 - d_2 P_0 + d_3 = 0
\]
where 
\[
d_1 = 3+2c_1+2c_2-4c_3, ~d_2 = 4c_1+4c_2 - 4c_3 + 7,~d_3 = (c_1+2)(c_2+2) - c_3
\]
and 
\[
P_0 = \frac{d_2 \pm \sqrt{d^2_2 - 4 d_1 d_3}}{2d_1}.
\]

\paragraph{Step 3} It is easy to show that $P'_0 = 1-P_0, ~\eta'_A = 1-\eta_A,~\eta'_B=1-\eta_B$ also satisfies the system of equations. But clearly this is not the true parameter set (violating assumption $\eta_A,~\eta_B<0.5$ ). 

\end{proof}

%\subsection*{Justification for the additional regret term in Section \ref{nogt}}
\subsection*{Proof for Theorem \ref{error:peer}}
%Denote the estimation of $\eta_A, \eta_B$ at time $t$ as $\hat{\eta}_{A,t}, \hat{\eta}_{B,t}$ respectively using estimates of $c_{1,t},c_{2,t},c_{3,t}$. A finer degree analysis also gives us:
%\begin{theorem}\label{error:peer}
%At $t$, w.p. $\geq 1-3\delta$, 
%$
%|\hat{\eta}_{A,t}-\eta_A | \leq O(\sqrt{\frac{\ln \frac{2}{\delta}}{2t}}), ~|\hat{\eta}_{B,t}-\eta_B| \leq O(\sqrt{\frac{\ln \frac{2}{\delta}}{2t}}),
%$ when $P_0$ is bounded away from 0.5.\footnote{When $P_0$ is close to 0.5, Eqn 1 and 2 can uniquely determine $\eta_A,\eta_B$ separately. }
%\end{theorem}
%This will again translates to a $O(\sqrt{\frac{\ln (6/\delta)}{2t}})$ regret bound for $\eta_A$ and $\eta_B$ with probability at least $1-\delta$, which incurs an additional $\sum_{t=1}^T O(\sqrt{\frac{\ln (6/\delta)}{2t}}) = O(\sqrt{T\cdot \ln \frac{6}{\delta}})$ regret, per Theorem~\ref{reg:error}. 

\begin{proof}
Define the following estimates:
\begin{align*}
&\hat{c}_{1,t} = \frac{\sum_{\tau=1}^t \mathbbm{1}(\hat{y}_{l,\tau} = 1)}{t},\\
&\hat{c}_{2,t} =  \frac{\sum_{\tau=1}^t \mathbbm{1}(\hat{y}_{r,\tau}=1)}{t},\\
&\hat{c}_{3,t} = \frac{\sum_{\tau=1}^t \mathbbm{1}(\hat{y}_{l,\tau} = \hat{y}_{r,\tau}=1)}{t}
\end{align*}
Using Hoeffding inequality we know that with probability at least $1-3\delta$
\[
|\hat{c}_{1,t} - c_{1,t} | \leq O(\sqrt{\frac{\ln (2/\delta)}{2t}}),~|\hat{c}_{2,t} - c_{2,t} | \leq O(\sqrt{\frac{\ln (2/\delta)}{2t}}) , ~|\hat{c}_{3,t} - c_{3,t} | \leq O(\sqrt{\frac{\ln (2/\delta)}{2t}}) 
\]
Then we prove that
\[
|\hat{P}_{0,t} - P_{0,t}| \leq  O(\sqrt{\frac{\ln (2/\delta)}{2t}}).
\]
when $t$ is sufficiently large. This can be proved easily via writing out the closed form solution for $P_0$ using Eqn. \ref{P0}:
\[
P_0 = \frac{d_2 \pm \sqrt{d^2_2 - 4 d_1 d_3}}{2d_1}
\], from which we know the error in estimating $P_0$ will be linear in errors in estimating $c_1,c_2,c_3$: this is because 
\[
2d_1 = 2(3+2c_1+2c_2-4c_3) = 6+4(c_1+c_2 -2 c_3) \geq 6.
\]
The inequality is due to $c_1 \geq c_3, c_2 \geq c_3$ by definition. Therefore when $t$ is large enough or the estimation errors in $c_1,c_2,c_3$ are small enough, the estimated $2\hat{d}_1$ will be bounded away from 0.

Since errors in estimating $\eta_A,\eta_B$ are also linear in the errors in $P_0$:
\[
\eta_A = \frac{c_1 - (1-P_0)}{2P_0-1}, ~\eta_B = \frac{c_2 - (1-P_0)}{2P_0-1},
\] this completes the proof.
\end{proof}

\subsection*{Proof for Theorem \ref{reg:hetero}}
\begin{proof}

Define 
\[
\hat{a}^* = \text{argmin}_i \sum_{t=1}^T  (1-2\tilde{\eta}_t) \ell_{i,t}
\]
First we have
\begin{align*}
 &\sum_{t=1}^T  \ell_{a,t} - \ell_{a^*,t} =  \sum_{t=1}^T  (\ell_{a,t} - \ell_{\hat{a}^*,t}) + \sum_{t=1}^T ( \ell_{\hat{a}^*,t} - \ell_{a^*,t})
\end{align*}
We define the following property of order-preserving:
\begin{align}\label{order:preserve}
 \sum_{t=1}^T \ell_{i,t} >  \sum_{t=1}^T \ell_{a^*,t} \Leftrightarrow \sum_{t=1}^T (1-2\tilde{\eta}_t)\ell_{i,t} >  \sum_{t=1}^T (1-2\tilde{\eta}_t) \ell_{a^*,t} 
 \end{align}
and we prove the following lemma:
\begin{lemma}\label{order:et}
For any $\alpha = 2+\epsilon, ~\epsilon > 0$, there exists a $\hat{p} < 1/2$ such that Eqn. \ref{order:preserve} holds for any agent $i$ whose accumulative regret satisfies that $L_i \leq (1+\alpha) L_{a^*}$.
\end{lemma}
The above implies that $\sum_{t=1}^T  \ell_{\hat{a}^*,t} - \ell_{a^*,t} \leq  \alpha L_{a^*}$. Now we will focus on $\sum_{t=1}^T  (\ell_{a,t} - \ell_{\hat{a}^*,t})$.  

Recall that $s_{i,t} = \ell(p_i(t), \tilde{y}_t) + \hat{\eta}_tp_i(t)(1-p_i(t))$. Via applying the martingale bound we established earlier in Section \ref{sec:peer}, with probability at least $1-\delta$ (via union bound)
\begin{align*}
&\sum_{t=1}^T  (1-2\tilde{\eta}_t) \cdot (\ell_{a,t} - \ell_{\hat{a}^*,t})  \\
\in &\sum_{t=1}^T  (1-2\tilde{\eta}_t)  \cdot (f(p_a(t),p_t) -f(p_{\hat{a}^*}(t),p_t)\pm \Et_{mart}(\frac{\delta}{2N}, 2, T)~~\text{($\ell$ is $f$ calibrated at each step $t$)}\\
\in &\sum_{t=1}^T  (s_{a,t} - s_{\hat{a}^*,t})\pm (\Et_{mart}(\frac{\delta}{2N}, 2, T)+\Et_{mart}(\frac{\delta}{2N}, \sigma_g, T)) \\
&~~~~+  2 \sum_{t=1}^T (\tilde{\eta}_t-\hat{\eta}_t)\biggl( p_a(t)(1-p_a(t)) - p_{\hat{a}^*}(t) (1-p_{\hat{a}^*}(t))\biggr)~~\text{($s$ is $1-2\tilde{\eta}_t$ compatible at each step $t$)}\\
\in & \sum_{t=1}^T  (s_{a,t} - s_{\hat{a}^*,t})\pm(\Et_{mart}(\frac{\delta}{2N}, 2, T)+\Et_{mart}(\frac{\delta}{2N}, \sigma_g, T))
\end{align*}
The above implies two things:
\begin{itemize}
\item With probability at least $1-\delta_g$, $\hat{a}^* = a^*_{peer}$.
\item $\sum_{t=1}^T  (1-2\tilde{\eta}_t) \cdot (\ell_{a,t} - \ell_{\hat{a}^*,t}) \leq \sum_{t=1}^T  (s_{a,t} - s_{\hat{a}^*,t})+(\Et_{mart}(\frac{\delta}{2N}, 2, T)+\Et_{mart}(\frac{\delta}{2N}, \sigma_g, T))
$
\end{itemize}
The above jointly implies that
\[
\sum_{t=1}^T  (1-2\tilde{\eta}_t) \cdot (\ell_{a,t} - \ell_{\hat{a}^*,t}) \leq \Et_{mart}(\frac{\delta}{2N}, 2, T)+\Et_{mart}(\frac{\delta}{2N}, \sigma_g, T)+\Et_{online}(T,N)
\]
Denote the RHS as $\Et_{total}$. Now we notice that 
\begin{align*}
& \sum_{t=1}^T  (1-2\tilde{\eta}_t) \cdot (\ell_{a,t} - \ell_{\hat{a}^*,t}) \geq (1-2\max_t \tilde{\eta}_t) \sum_{t=1}^T \ell_{a,t} -  \sum_{t=1}^T \ell_{\hat{a}^*,t}
\end{align*}
This further implies 
\begin{align*}
 &(1-2\max_t \tilde{\eta}_t) (\sum_{t=1}^T \ell_{a,t} - \sum_{t=1}^T \ell_{\hat{a}^*,t})\\
 \leq &   \sum_{t=1}^T  (1-2\tilde{\eta}_t) \cdot (\ell_{a,t} - \ell_{\hat{a}^*,t})  + 2\max_t \tilde{\eta}_t \sum_{t=1}^T \ell_{\hat{a}^*,t}\\
 \leq & \Et_{total} + L_{\hat{a}^*} \\
 \leq & \Et_{total} + \alpha L_{a^*}
\end{align*}
Therefore 
\[
\sum_{t=1}^T \ell_{a,t} - \sum_{t=1}^T \ell_{\hat{a}^*,t} \leq \frac{\Et_{total} + \alpha L_{a^*}}{1-2\max_t \tilde{\eta}_t}
\]
Putting everything up together, we prove that with probability at least $1-\delta-\delta_g$
\begin{align*}
& \sum_{t=1}^T  \ell_{a,t} - \ell_{a^*,t} \\
= & \sum_{t=1}^T  (\ell_{a,t} - \ell_{\hat{a}^*,t}) + \sum_{t=1}^T ( \ell_{\hat{a}^*,t} - \ell_{a^*,t})\\
\leq &\frac{\Et_{total} + \alpha L_{a^*}}{1-2\max_t \tilde{\eta}_t} + \alpha L_{a^*}\\
\leq & \frac{\Et_{mart}(\frac{\delta}{2N}, 2, T)+\Et_{mart}(\frac{\delta}{2N}, \sigma_g, T)+\Et_{online}(T,N)}{1-2\max_t \tilde{\eta}_t} + \alpha\left(\frac{1}{1-2\max_t \tilde{\eta}_t}+1\right)L_{a^*}
\end{align*}
\end{proof}

\subsection*{Proof for Lemma \ref{order:et}}
\begin{proof}
Recall $\Delta_i = \sum_{t=1}^T \ell_{i,t} - \sum_{t=1}^T \ell_{a^{*},t}$. And we know 
\[
(1-2\tilde{\eta}) \cdot (\sum_{t=1}^T \ell_{i,t} - \sum_{t=1}^T \ell_{a^*,t}) = (1-2\tilde{\eta}) \cdot \Delta_i
\]
Observe that
\begin{align*}
&\sum_{t=1}^T  (1-2\tilde{\eta}_t) \cdot (\ell_{i,t} - \ell_{a^*,t})- 
(1-2\tilde{\eta}) \cdot (\sum_{t=1}^T \ell_{i,t} - \sum_{t=1}^T \ell_{a^*,t}) = \sum_{t=1}^T  2(\tilde{\eta} - \tilde{\eta}_t) \cdot (\ell_{i,t} - \ell_{a^*,t})
\end{align*}
Denote 
\begin{align*}
\mathcal S_{i,+} &:= \{t: \ell_{i,t} - \ell_{a^*,t} \geq 0\}, ~
\mathcal S_{i,-} = \{t: \ell_{i,t} - \ell_{a^*,t} <  0\}\\
\ell_{i,+}  &:= \sum_{t \in \mathcal S_{i,+}} \ell_{i,t} - \ell_{a^*,t},~\ell_{i,-}  := \sum_{t \in \mathcal S_{i,-}} \ell_{i,t} - \ell_{a^*,t},~
\end{align*}
Notice $\ell_{i,-} < 0$. Denote $\epsilon_{\eta} := \max_t |\eta_t - \tilde{\eta}|$. Then we have
\[
\sum_{t=1}^T 2 (\tilde{\eta} - \tilde{\eta}_t) \cdot (\ell_{i,t} - \ell_{a^*,t}) \leq 2\epsilon_{\eta} \cdot (\ell_{i,+} - \ell_{i,-})
\]
Denote the number of times $i$ disagrees with $a^*$ as $J$. Then we have
\[
\sum_{t=1}^T \ell_{i,t} \geq J - L_{a^*} \Rightarrow J \leq \sum_{t=1}^T \ell_{i,t} + L_{a^*} = 2L_{a^*} + \Delta_i
\]
Further
\[
\ell_{i,+} - \ell_{i,-} = J \leq 2L_{a^*} + \Delta_i \Leftrightarrow 2\epsilon_{\eta}(\ell_{i,+} - \ell_{i,-}) \leq 2\epsilon_{\eta} (2L_{a^*} + \Delta_i) \leq (2+\alpha) L_{a^*}
\]
To summarize,
\begin{align*}
&\sum_{t=1}^T  (1-2\tilde{\eta}_t) \cdot (\ell_{i,t} - \ell_{a^*,t})- 
(1-2\tilde{\eta}) \cdot (\sum_{t=1}^T \ell_{i,t} - \sum_{t=1}^T \ell_{a^*,t}) \\
\leq &2\epsilon_{\eta}(\ell_{i,+} - \ell_{i,-}) \\
\leq &2\epsilon_{\eta} (2L_{a^*} + \Delta_i)
\end{align*}
As long as above difference bound $2\epsilon_{\eta} (2L_{a^*} + \Delta_i)$ is smaller than $(1-2\tilde{\eta})\cdot  \Delta_i$, i.e., 
\[
2\epsilon_{\eta} \leq (1-2\tilde{\eta}) \cdot \frac{\Delta_i}{2L_{a^*}+\Delta_i}
\]
we will not flip the order of $i$ and $a^*$. 
If we allow selecting an agent within $(1+\alpha)L_{a^*}$, then $ \Delta_i \leq \alpha L_{a^*}$, and
\[
(1-2\tilde{\eta}) \cdot \frac{\Delta_i}{2L_{a^*}+\Delta_i} \geq (1-2\tilde{\eta})\cdot \frac{\alpha}{2+\alpha} > \frac{1-2\tilde{\eta}}{2},~\forall \alpha > 2.
\]
The rest to show is there exists $\hat{p}$ that admits 
\[
 (1-2\tilde{\eta})\cdot \frac{\alpha}{2+\alpha} \leq 2\epsilon_{\eta} \Leftarrow \frac{1}{2} < \tilde{\eta} + \epsilon_{\eta}
\]
Denote $\bar{\eta} := \frac{\sum_{t=1}^T \eta_t}{T}$, the average error rates before flipping. Note 
\[
\tilde{\eta}  = \bar{\eta} (1-2\hat{p}) + \hat{p} > \hat{p}
\]
where remember $\bar{\eta}$ is the average error rates before random flipping. This leads to
\begin{align}
\tilde{\eta}+ \epsilon_{\eta}  > \hat{p} + \epsilon_{\eta}.
\end{align}
Therefore as long as $\hat{p} \geq \frac{1}{2} - \epsilon_{\eta}$, the condition will be satisfied.
\end{proof}